\documentclass{ecai} 

\usepackage{latexsym}
\usepackage{amssymb}
\usepackage{amsmath}
\usepackage{amsthm}
\usepackage{booktabs}
\usepackage{enumitem}
\usepackage{graphicx}
\usepackage[table]{xcolor}
\usepackage{xcolor}
\usepackage{url}
\usepackage{subcaption}
\usepackage{multirow}
\usepackage{array}
\usepackage{hyperref}
\usepackage{fancyhdr}
\fancypagestyle{firstpage}{
    \fancyhf{} 
    \fancyhead[C]{Published as a conference paper in ECAI 2025}
    
}

\newcommand{\BibTeX}{B\kern-.05em{\sc i\kern-.025em b}\kern-.08em\TeX}

\begin{document}


\begin{frontmatter}


\paperid{123} 


\title{Efficient Context Propagating Perceiver Architectures for Auto-Regressive Language Modeling}

\author[A]{\fnms{Kaleel}~\snm{Mahmood}\thanks{Corresponding author Email: kaleel.mahmood@uri.edu}}
\author[B]{\fnms{Shaoyi}~\snm{Huang}}
\address[A]{Department of Computer Science and Statistics, University of Rhode Island, Kingston, RI, USA}
\address[B]{Department of Computer Science, Stevens Institute of Technology, Hoboken, New Jersey, USA}


\begin{abstract}
One of the key challenges in Transformer architectures is the quadratic complexity of the attention mechanism, which limits the efficient processing of long sequences. Many recent research works have attempted to provide a reduction from the $O(n^2)$ time complexity of attention to semi-linear complexity. However, it remains an unsolved problem in the sense of maintaining high performance when complexity is reduced. One of the important works in this respect is the Perceiver class of architectures that have demonstrated excellent performance, while reducing the computation complexity. In this paper, we use the PerceiverAR as a basis and explore the design space of different trade-offs between preserving context and reducing attention complexity. To this end, we develop four new architectural paradigms, the best performing of which we denote as the Efficient Context propagating Perceiver (ECP). ECP has two major advantages over the PerceiverAR. First,  the ECP architecture overcomes the main drawback of PercieverAR by utilizing both the context and the latent sequences in autoregressive training. Second, the ECP architecture operates with the same attention complexity as LongLoRA, making it computationally efficient. More importantly, via pairwise segment attention, it extracts better information resulting in improved language modeling. Empirically, we demonstrate that the ECP architecture significantly outperforms other state-of-the-art Transformer models on Wikitext-103, PG-19 and sCIFAR-10.

\noindent\textbf{Code available at:}  \url{https://github.com/MetaMain/ECPTransformer}

\end{abstract}

\end{frontmatter}

\thispagestyle{firstpage}
\section{Introduction}
\label{sec:intro}
The Transformer architecture has revolutionized the field of artificial intelligence, especially in Natural Language Processing (NLP)~\cite{vaswani2017attention}. The recent success of Large Language Models (LLMs) such as ChatGPT ~\cite{achiam2023gpt}, Gemini~\cite{team2023gemini}, Llama~\cite{touvron2023llama, dubey2024llama}, etc. with their comprehension and reasoning capabilities, is a testament to the effectiveness of the Transformer architecture. Prior to Transformers, deep Convolutional Neural Networks (CNNs) had demonstrated impressive results in computer vision applications, however, their performance does not show the same effectiveness when applied to NLP. One of the reasons CNNs are typically less effective in NLP is their limited receptive field, due in part to the convolution operation. The Transformer on the other hand, uses the attention mechanism. This operation measures the pairwise similarity between the words or tokens of the entire input sequence in order to comprehend it.

\begin{figure*}[!ht]
  \centering

     \includegraphics[width=.90\textwidth]{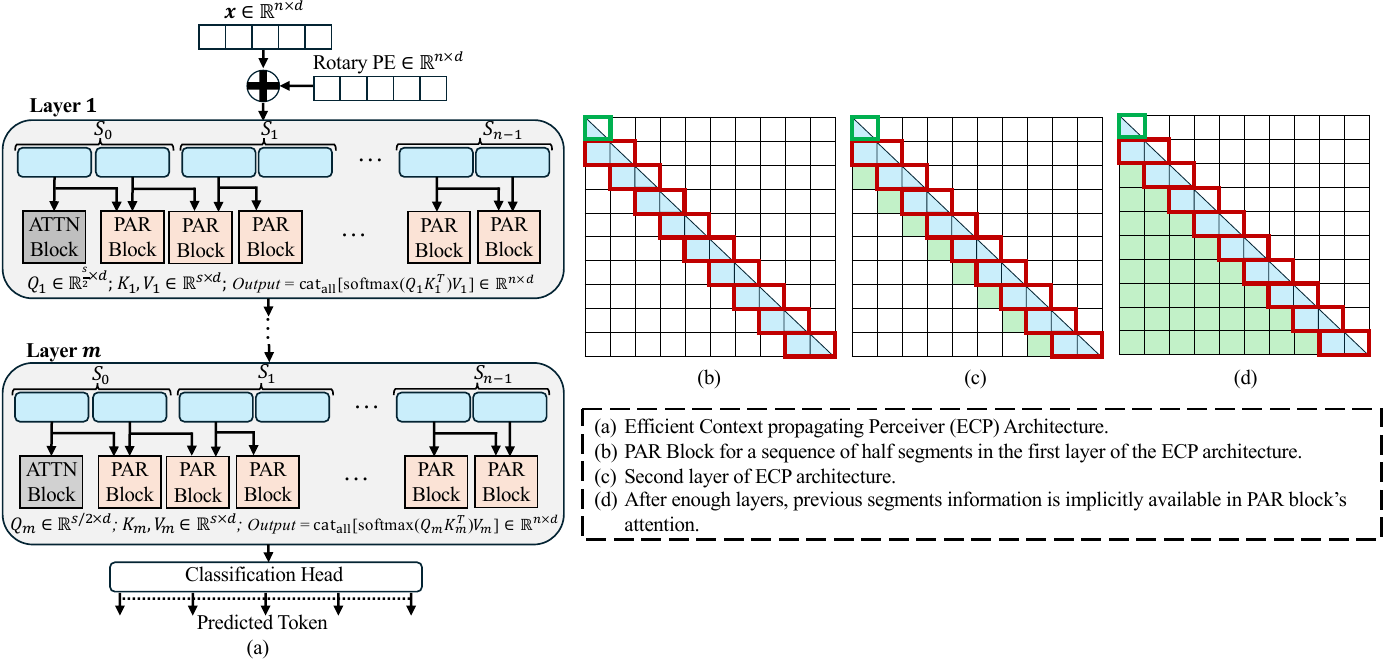}
    \caption{Subfigure (a) shows the ECP architecture. In subfigures (b)-(d) the green rectangles indicate the propagation of attention, the red rectangles indicate the PerceiverAR pairwise segment attention. The green blocks are not calculated but contain information because of the PAR block in the previous layer. The attention calculation is done only on the two blocks near the diagonal (indicated by red rectangles).}
    \label{fig:LLP}
     \centering
\end{figure*}

Attention is one of the major computational operations in Transformer architectures with an $O(n^2)$ complexity. There has been a great deal of research to reduce the complexity of attention while maintaining high performance. Of particular note is the PerceiverAR~\cite{hawthorne2022general}. The PerceiverAR  efficiently handles long contexts by dividing the input sequence in to two components, history (or the context) and latent. However, in this architecture, the history component of the input is not used in autoregressive learning. An additional drawback of the PerceiverAR is the history component gets compressed into the latent part after the first layer. In this work, we design a new architecture called Efficient Context Propagating Perceiver (ECP). The ECP architecture uses local pairwise segment attention making it more computationally efficient than the original PerceiverAR. More importantly, this leads to better learning as we demonstrate in later sections. The Efficient Context Propagating Perceiver is shown in Figure~\ref{fig:LLP}(a) and the visual depiction of the associated algorithm is shown in Figures~\ref{fig:LLP}(b-d). As can be seen in the figures, the local pairwise segment attention gets propagated to the entire context as more layers are used in the Transformer. We provide more details on this design in later sections. Overall, our work advances the field of autoregressive language modeling and provides the following contributions:
\begin{enumerate}
    \item \textbf{Design of Context Efficient Architectures} - We design three different architectures to overcome the loss of information inherent in the  history or context part of the PerceiverAR, as it is merged into the latent after the first layer. Each architecture has computation efficiency  versus performance trade-offs which we analyze. 
    
    \item \textbf{Efficient Context Propagating PerceiverAR (ECP)} - We develop a novel segment attention algorithm that is highly efficient and has similar complexity as LongLoRA. Simultaneously, it achieves implicitly full attention while computing attention only on pairs of overlapping segments in a PerceiverAR style.

 \item \textbf{Empirical Evaluation} - Experimental results show that our proposed ECP architecture outperforms SOTA models on various datasets. On Wikitext-103, and PG-19 benchmarks, ECP achieves significantly lower perplexity than equivalent size SOTA models. 
  
\end{enumerate}

\section{Background and Related Work}
\label{sec:RelatedWork}

For a Transfomer, consider the matrix $Q$, $K$ and $V$ containing rows representing the learnt position encoded (PE) embedding of each token in $d$ dimensions (i.e., $1 \times d$). The attention $A=\text{softmax}(QK^{T})$ contains the dot product similarity of each input token with every other token in the input sequence. For an input sequence with $n$ tokens, $Q,K \in \mathbb{R}^{n \times d}$ and attention $A \in \mathbb{R}^{n \times n}$. 

The Transformer divides the attention calculation into parallel heads so as to refine the learning, and better comprehend the contextual meaning of the input sequence. Each head computes the attention on a portion of the embedding dimension. The output in each head is computed by further multiplying the attention $A$ with $V$. The canonical Transformer’s operation can be summarized by the following equations~\cite{vaswani2017attention}. First the output of the $i^{th}$ head is:
\begin{equation}
    H_{i} = \text{softmax}(\frac{Q_{i}K^{T}_{i}}{\sqrt{d_{k}}})V_{i}=A_{i}V_{i}
\end{equation}
where $d_{k}=\frac{d}{h}$ is the dimension of each head, $h$ is the number of heads in each layer and $H_{i} \in \mathbb{R}^{n \times d_{k}}$.
The output in each Transformer layer $Z$, is obtained by catenating the output of all heads and transformed further by a projection matrix $W^{o}$:
\begin{equation}
    Z_{j} = \text{catenate}(H_{0},H_{1},...,H_{h-1})W^{o}
\end{equation}
where $W^{o} \in \mathbb{R}^{d \times d}$ and $Z{j} \in \mathbb{R}^{n \times d}$ (the same dimensions as the input). A classification layer is added to the last layer which predicts the next token in autoregressive generation:
\begin{multline}
  out = classification(Z_{p-1}(Z_{p-2}(...\\
  Z_{0}(embedding(x)+PE(x)))))
\end{multline}
Skip connections and layer normalization are also used in each layer to stabilize the training of the Transformer.

For language models, if text generation is the goal, the model is trained in an autoregressive manner, where it learns to predict the next token given an input sequence of tokens. In autoregressive generation, the previously predicted token becomes part of the next input sequence.  For NLP models, the training process can be made highly effective by masking the attention matrix in a triangular fashion so that future tokens are not visible. This helps in creating more (input, output) training pairs. From an input training sequence of size $n$, $n-1$ training pairs can be created by simply hiding the next token, one at a time. For NLP classification, the masking of the attention is not needed, as the classification decision is made on the entire input sequence. 

Since the attention computation in each head measures the pairwise similarity in the input sequence, its time complexity is $O(n^2)$ if the sequence length is $n$. With larger NLP models being created operating on longer sequence lengths, with multiple heads in each layer and many layers, the computational costs of Transformer training is becoming an important issue. Considerable research effort is being put into making the attention mechanism more efficient since it is the dominant computation in a transformer. Numerous research papers have proposed ideas to reduce the quadratic time complexity of attention to linear or sub quadratic complexity. 

Some of the important works in this respect include: TransformerXL~\cite{dai-etal-2019-transformer}, Linformer~\cite{wang2020linformer}, Longformer~\cite{beltagy2020longformer}, Reformer~\cite{kitaev2020reformer}, Performer~\cite{choromanski2020rethinking}, Long-Short Attention~\cite{zhu2021long}, and Perceiver~\cite{hawthorne2022general,jaegle2021perceiver,jaegleperceiver}, among others. Recently State Space Models~\cite{gu2021efficiently, fuhungry, daotransformers} have drawn considerable research attention producing impressive results in some domains. It is yet to be seen if they can be better alternatives for NLP generative models. Thus, we limit our comparisons to transformer-based architectures in this work. Next we elaborate on the PerceiverAR~\cite{hawthorne2022general} architecture  that we use as a basis to further build and arrive at an efficient and better transformer architecture.

\section{Preliminaries - PerceiverAR Architecture}
\label{sec:PerceiverArch}

The fundamental concept behind PerceiverAR~\cite{hawthorne2022general} is to split the input sequence into two components which we term as the history and latent sequences. We denote the input as $x$ with corresponding sequence length $n$. After the tokenization and embedding is carried out, the input can be considered as composed of a history component and a latent component as: $x \in \mathbb{R}^{n \times d} = x_{history} || x_{latent}$, where  $||$ indicates the catenation of two components. We denote the history length as $h$ and the latent length as  $l$. After the embedding operation, $history \in \mathbb{R}^{h \times d}$ and $latent \in \mathbb{R}^{l \times d}$. The PerceiverAR computes the query only on the latent part in the first layer of Transformer, while the key and values are computed on the entire sequence length of size $n$. Thus the attention computation in the first layer produces an output of dimension $\mathbb{R}^{l \times d}$:
\begin{equation}
\label{eq:perOne}
    Q_{latent} = W_{q}x_{latent} \in \mathbb{R}^{l \times d}
\end{equation}
\begin{equation}
\label{eq:perTwo}
    K = W_{k}x \in \mathbb{R}^{n \times d}
\end{equation}
\begin{equation}
    V = W_{v}x \in \mathbb{R}^{n \times d}
\end{equation}
\begin{equation}
\label{eq:perlast}
    Output = [\text{softmax}(Q_{latent}K^{T})V] = [AV]\in \mathbb{R}^{l \times d} 
\end{equation}

Since the output from first layer of Transformer $\in \mathbb{R}^{l \times d}$ the remaining layers do a normal attention on inputs of size $l$, without splitting the input into two parts as is done in the first layer. For autoregressive training, an input of $n$ tokens is used to create $(n-1)$ training pairs, such that the expected output of first token is the second token, and the expected token of inputs up to $(m-1)$ tokens is the $m^{th}$ token. Since PerceiverAR uses the history part as a fixed input, the autoregressive training can only be done on the latent part of input. Thus, to hide the future tokens in the training of the first layer, the upper triangular part of the attention matrix (corresponding to latent) is set to $-\infty$. The remaining layers operate on the size of the latent length, so the triangularization of the attention matrix is done on the entire square matrix. 

The attention complexity of PerceiverAR in the first layer of Transformer is $O(l \times n)$ while the remaining layers have a complexity of $O(l^{2})$. This provides a significant reduction in computation, especially when $l$ is much less than $n$ and many layers are used in the Transformer. While the PerceiverAR has been able to accomplish very good results on NLP benchmarks~\cite{hawthorne2022general}, it has two main drawbacks which we significantly improve in this work, finally leading to an impressive new architecture:
\begin{enumerate}
    \item \textbf{Latent Training Dependency} - The training for AutoRegressive generation can only use the latent part of the input. Therefore, more training is required to accomplish the same learning as a normal Transformer (provided the model does not overfit). 
    \item \textbf{Lossy History} - The history is implicitly compressed into the latent output of first layer and is not explicitly refined as in a normal Transformer in following layers.
\end{enumerate}
We improve upon the above drawbacks and present three different enhancements for better utilization of the history component in the PerceiverAR. This further leads to the best architecture developed in this work.

\section{New Architecture Paradigms for Latent History Attention and Efficiency}
\label{sec:ArchParadigm}

\begin{figure}
\centering
\subfloat[Baseline PerceiverAR]
{\includegraphics[clip,width=.5\linewidth]{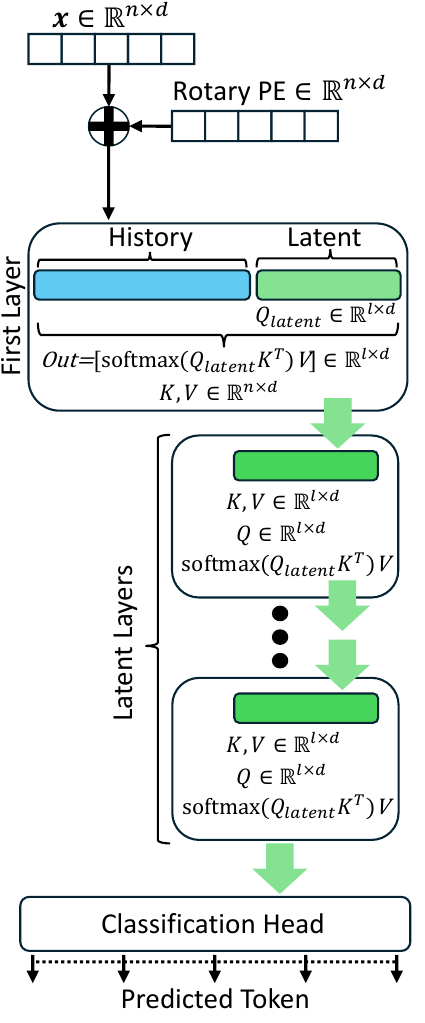}
\label{fig:Base}
}
\subfloat[Double Attention PerceiverAR]
{\includegraphics[clip,width=.5\linewidth]{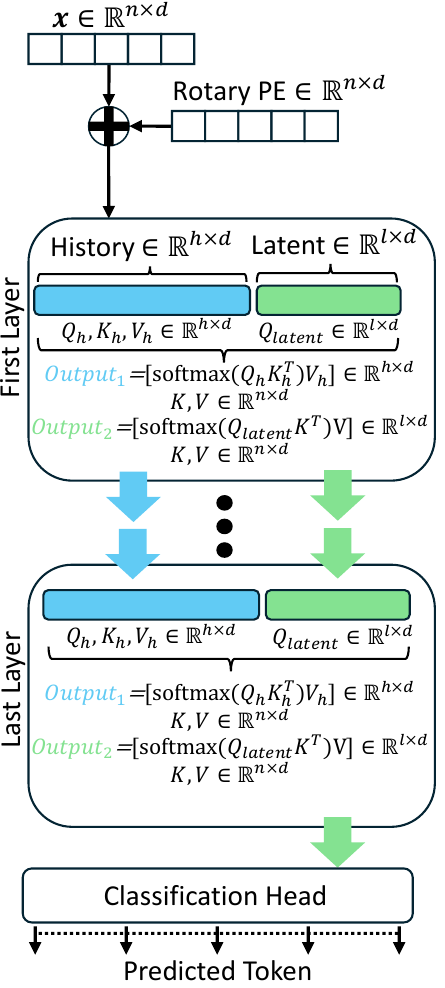}
\vspace{0.05in}
\label{fig:V1}
}
\vspace{0.1in}
\caption{Baseline~\cite{hawthorne2022general} and Double Attention PerceiverAR architectures.}
\vspace{0.1in}
\end{figure}

The baseline PerceiverAR (as shown in Figure~\ref{fig:Base}) uses the history information explicitly only in the first layer by computing the key and values on the entire input sequence, while the query is computed only on the latent part of the input. To overcome the loss of history information in subsequent layers, we propose the first architectural paradigm where each layer computes two attentions and correspondingly two outputs.

\subsection{Double Attention PerceiverAR Architecture}

In this enhancement (shown in Figure~\ref{fig:V1}), each layer performs two attention operations. The first attention is the same computation as the PerceiverAR baseline as given by Equations \ref{eq:perOne} through \ref{eq:perlast}. The second attention computation is based on the history component of the input, and is computed in each layer. The second attention generates an additional output as follows:
\begin{equation}
    Q_{h} = W_{qh}x_{history} \in \mathbb{R}^{h \times d}
\end{equation}
\begin{equation}
    K_{h} = W_{kh}x_{history} \in \mathbb{R}^{h \times d}
\end{equation}

\begin{equation}
    V_{h} = W_{kv}x_{history} \in \mathbb{R}^{h \times d}
\end{equation}
where $x_{history}$ is the history component of the input.
\begin{equation}
    Output_{1} = [\text{softmax}(Q_{h}K_{h}^{T})V_{h}] = [A_{h}V_{h}]\in \mathbb{R}^{h \times d} 
\end{equation}
\begin{equation}
    Output_{2} = [\text{softmax}(Q_{latent}K^{T})V] = [A_{latent}V]\in \mathbb{R}^{l \times d} 
\end{equation}
Thus each layer in the double attention PerceiverAR architecture is identical. The two outputs corresponding to the latent attention and the history attention are concatenated to become the single output, and therefore the corresponding input for the subsequent layer. Note that no masking is used in the attention on the history part, as this part is not used for autoregressive training. Only the latent attention uses triangular masking. 

The overhead in this architecture is the computation of the attention in the history component of the input. If the history length $h$ is larger than the latent length $l$, then this could be significantly more computations as compared to the baseline PerceiverAR where the subsequent layers after the first layer compute attention only on the latent part. To improve this drawback and reduce the computational overhead, we propose a second architectural paradigm as described in the next subsection.

\subsection{Compressed Double Attention PerceiverAR Architecture}
\begin{figure}
  \centering
    \includegraphics[width=0.5\linewidth]{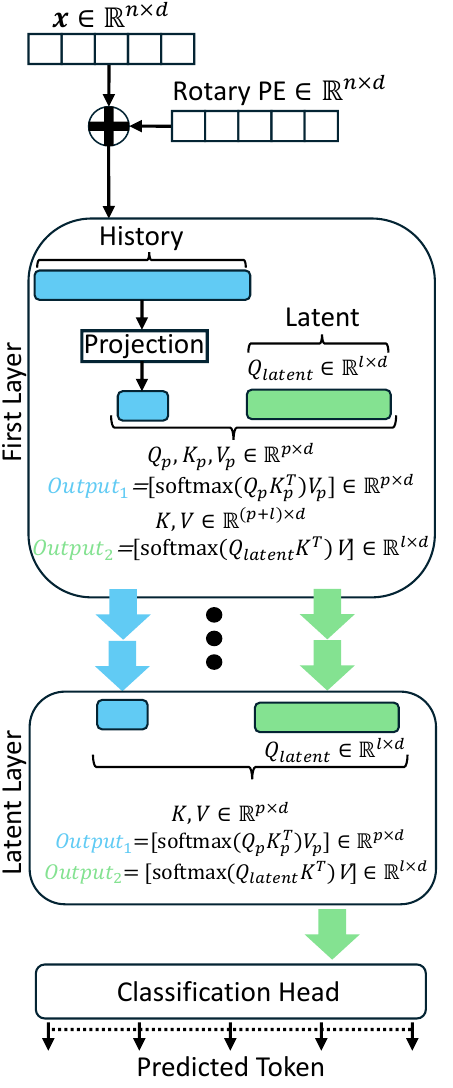}
    \caption{Compressed Double Attention PerceiverAR,}
    \label{fig:V2}
    \vspace{0.3in}
\end{figure}
To improve the efficiency of the history attention computation, we refine the double attention PerceiverAR architecture. We reduce the attention overhead by having the first layer compress the history part of the input by projecting it to a smaller length along the sequence dimension as shown in Figure \ref{fig:V2}. This compresses the history information and this compressed history is then used and refined in all remaining layers. We detail the mathematics behind this architecture as follows: First, the projection of the history to a compressed length $p$ is carried out only in the first layer:
\begin{equation}
x_{ph} = W_{ph}x_{history} \in \mathbb{R}^{p \times d}  
\end{equation}
All layers including the first layer implement the following: 
\begin{equation}
    Q_{ph} = W_{qh}x_{ph} \in \mathbb{R}^{p \times d}
\end{equation}
\begin{equation}
K_{ph} = W_{kh}x_{ph} \in \mathbb{R}^{p \times d}
\end{equation}
\begin{equation}
V_{ph} = W_{kv}x_{ph} \in \mathbb{R}^{p \times d}
\end{equation}
\begin{equation}
    Q_{latent} = W_{ql}x_{latent} \in \mathbb{R}^{l \times d}
\end{equation}
\begin{equation}
K = W_{k}(x_{ph}||x_{latent}) \in \mathbb{R}^{(p+l) \times d}
\end{equation}
\begin{equation}
V = W_{v}(x_{ph}||x_{latent}) \in \mathbb{R}^{(p+l) \times d}
\end{equation}
\begin{equation}
    Output_{1} = [\text{softmax}(Q_{ph}K_{ph}^{T})V_{ph}] = [A_{ph}V_{h}]\in \mathbb{R}^{p \times d} 
\end{equation}
\begin{equation}
    Output_{2} = [\text{softmax}(Q_{latent}K^{T})V] = [A_{latent}V]\in \mathbb{R}^{l \times d} 
\end{equation}

It is important to note that the compressed double attention architecture is one possible way to achieve efficient attention computations. While a variety of other approaches are possible, our next architectural paradigm demonstrates another intuitive way that efficient computation can be achieved.

 \subsection{$s$-Split Double Attention PerceiverAR}
\begin{figure}
  \centering
    \includegraphics[width=0.5\linewidth]{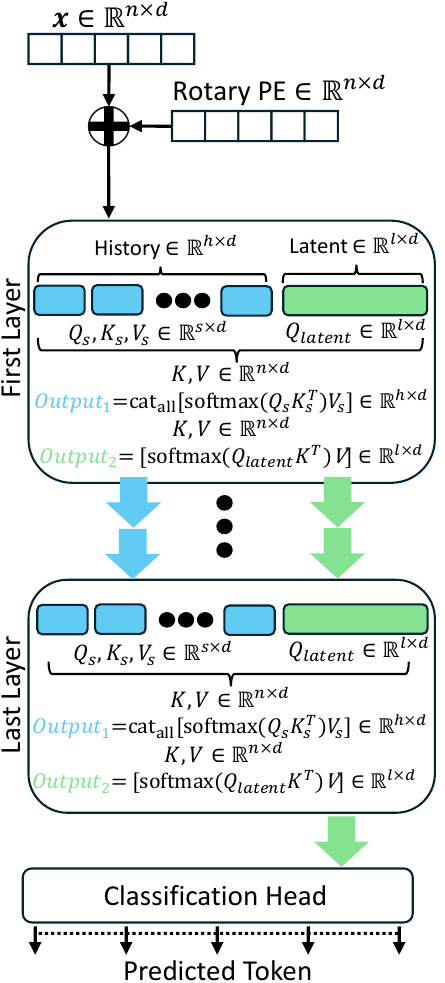}
    \caption{$s$-Split Double Attention PerceiverAR.}
    \label{fig:V3}
    \vspace{0.2in}
\end{figure}

 In this architectural paradigm, efficient computation is achieved by dividing the history component into small segments of size $s$. If the history segment size $s$ is smaller than the latent length $l$, i.e., $s <<  l$, then the overhead in the history computation in each layer is minimal. Note that the segment-wise attention is carried out within the same segment only, so if the segment size is $s$, the complexity of attention in each segment is $O(s^{2})$. The output corresponding to each segment is catenated to become the history component of the output. A visualization for the $s$-Split double attention PerceiverAR is shown in Figure~\ref{fig:V3}.
 
The above architecture paradigm leads to a highly efficient information propagation and extraction paradigm that we detail in the next section.
\section{Efficient Context Propagating Perceiver (ECP) Architecture}
\label{sec:ECP}
In general, the complexity of computing attention is $O(n^2)$. In Section~\ref{sec:ArchParadigm}, we introduced several novel paradigms to propagate the history component more effectively in the PerceiverAR architecture, and also make the history attention computationally efficient. First, in the double attention PerceiverAR, the attention is computed for both the history and latent component. The drawback here is the added computational burden of computing double the number of attentions. One way to reduce this complexity is by dividing the input sequence into disjoint segments, and only computing the attention in each segment itself (the $s$-split double attention PerceiverAR). However, the disadvantage of such an approach is that there is some loss of context due to lack of information flow between disjoint segments. Likewise, the compressed double attention PerceiverAR may also lose information as the history component is compressed in the first layer.

One way that attention can be efficiently computed is with LongLoRA~\cite{chenlonglora} which solves this problems via shifted sparse attention ($S^{2}$ Attn). In $S^{2}$ Attn, the sequence length is split into different groups and each group computes the attention individually. To support the information flow between different groups, the attention heads are divided in two halves. In the second half of the groups, the tokens are shifted by half the group size. This simple shift causes the information to be shared between neighboring groups. The primary application of LongLoRA was demonstrated in applying LoRA (Low Rank Adaptation) to the self attention layers in extending the sequence length of existing LLMs. 

\begin{figure}
  \centering
    \includegraphics[width=0.7\linewidth]{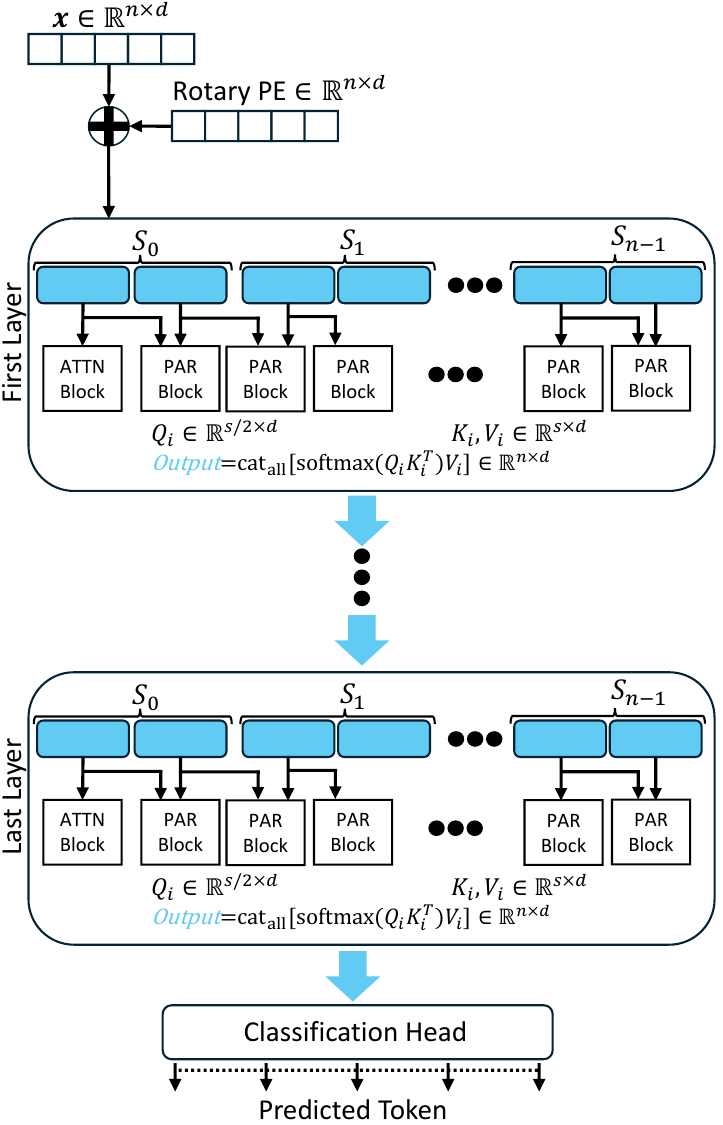}
    \caption{Efficient Context propagating Perceiver
(ECP)}
    \label{fig:LLP_2}
    \vspace{0.2in}
\end{figure}

Inspired by the idea of LongLoRA’s overlapping attention, we apply this concept to enhance the PerceiverAR design by dividing the input sequence into segments, such that the PerceiverAR allows communication of information from previous segment to the current segment. This is accomplished by first dividing the input sequence into disjoint segments. Next each segment is further divided into two halves, a history component and a latent component. The standard PerceiverAR computation is applied to consecutive pairs of half segments. Specifically, $Q$ is computed on the current half segment while $K$ and $V$ are computed on the current and previous half segments. Thus, the PerceiverAR  operates upon overlapping segments which allows information flow down the layers. We denote this architecture as the Efficient Context propagating Perceiver (ECP). It is depicted in Figure~\ref{fig:LLP_2}.

The ATTN block in Figure \ref{fig:LLP_2} is a regular attention block which operates upon half of the first segment only. The attention equations governing the ATTN block with segment size $s$ are:
\begin{equation}
    Q_{atn} = W_{qatn}x_{0:\frac{s}{2}} \in \mathbb{R}^{\frac{s}{2} \times d}
\end{equation}
\begin{equation}
    K_{atn} = W_{katn}x_{0:\frac{s}{2}} \in \mathbb{R}^{\frac{s}{2} \times d}
\end{equation}
\begin{equation}
    V_{atn} = W_{vatn}x_{0:\frac{s}{2}} \in \mathbb{R}^{\frac{s}{2} \times d}
\end{equation}
\begin{multline}
        Output_{atn} = [\text{softmax}(Q_{atn}K^{T}_{atn})V_{atn}] \\= [A_{atn}V_{atn}]\in \mathbb{R}^{\frac{s}{2} \times d} 
\end{multline}
The PAR block performs the PerceiverAR operation on two consecutive half segments. The operations of the $i^{th}$ PAR block is given by the following set of equations: 
\begin{equation}
\label{eq:firstV4}
Q_{par_{i}} = W_{qpart_{i}}x_{(\frac{s}{2})(i):(\frac{s}{2})(i+1)} \in \mathbb{R}^{\frac{s}{2} \times d}    
\end{equation}
\begin{equation}
K_{par_{i}} = W_{kpart_{i}}x_{(\frac{s}{2})(i-1):(\frac{s}{2})(i+1)} \in \mathbb{R}^{s \times d}     
\end{equation}
\begin{equation}
V_{par_{i}} = W_{vpart_{i}}x_{(\frac{s}{2})(i-1):(\frac{s}{2})(i+1)} \in \mathbb{R}^{s \times d}      
\end{equation}
\begin{multline}
Output_{par_{i}} = [\text{softmax}(Q_{par_{i}}K^{T}_{par_{i}})V_{par_{i}}] \\= [A_{par_{i}}V_{par_{i}}]\in \mathbb{R}^{\frac{s}{2} \times d} 
\label{eq:lastV4}
\end{multline}
where in Equations~\ref{eq:firstV4} to~\ref{eq:lastV4}, $i \in (1,2\dots s_{total})$ and $s_{total}$ is the total number of segments. Each of the PAR blocks outputs data equal to half of the segment size i.e., $\mathbb{R}^{\frac{s}{2} \times d}$. All blocks use $Q \in \mathbb{R}^{\frac{s}{2} \times d}$ while $K, V \in \mathbb{R}^{s \times d}$ are computed on double the size i.e., on the current half segment and the previous half segment. Thus to implement autoregressive behavior, masking is done on the right $\frac{s}{2} \times \frac{s}{2}$ part of the attention in PAR block. Note that PAR attention is of size $\frac{s}{2}\times s$. The very first block in each layer is different and does a normal attention computation with triangular masking on the first half segment. This is done so that even the first half segment can be used in autoregressive modeling, unlike a normal PerceiverAR where the history part cannot be used in autoregressive training in terms of masking this part. All blocks in a layer output data of $\frac{s}{2}\times d$ size. These are then concatenated to form an output of size $n \times d$. All layers in the ECP architecture are identical as shown in Figure~\ref{fig:LLP_2}.

\subsection{ECP Analysis and Advantages}

\begin{figure}
  \centering
    \includegraphics[width=1\linewidth]{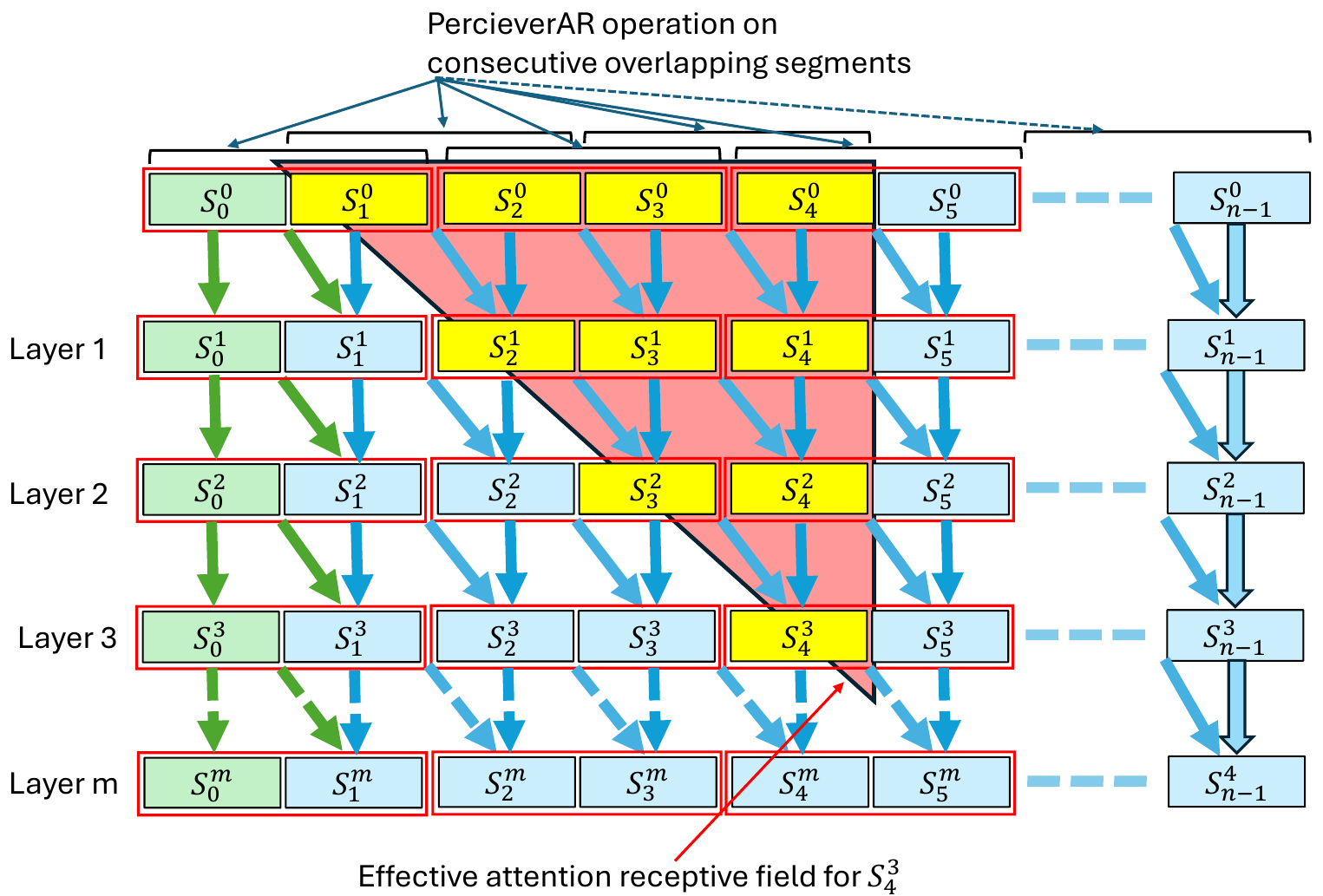}
    \caption{Visual depiction of the ECP Algorithm indicating increase in the attention receptive field as computation proceeds down the layers of Transformer.}
    \label{fig:LLPExp}
    \centering
    \vspace{0.2in}
\end{figure}

In the ECP algorithm, the PerceiverAR operation is performed on a pair of overlapping half segments $S_{i}^{j}$, where $i$ is the half segment number and $j$ is the layer number in each layer of Transformer. This is visually depicted in Figure~\ref{fig:LLPExp}. The effective receptive attention field increases as computation progresses down the layers of Transformer. For example, the calculation of the output in segment $S4$  in layer 3 is denoted as $S_{4}^{3}$ and uses $S_{4}^{2}$ and $S_{3}^{2}$ from the previous layer (layer 2). $Q$ is computed on $S_{4}^{2}$ while $K$ and $V$ are computed on both $S_{3}^{2}$ and $S_{4}^{2}$. Note that $S_{3}^{2}$ in turn uses $S_{2}^{1}$ and $S_{3}^{1}$ from the previous layer (layer 1), and $S_{2}^{1}$ further uses $S_{1}^{0}$ and $S_{2}^{0}$. The segment information accumulated by $S_{4}^{3}$ from previous segments is depicted with a yellow color in Figure~\ref{fig:LLPExp}. Thus, even though the PerceiverAR style attention computation is local with size $\frac{s}{2} \times s$ if the segment size is $s$, the propagation of information from all previous segments occurs down the layers. The first half segment is treated as a special case (as it cannot form a pair with a previous segment), and normal full attention is carried out on it with upper triangular masking, to aid in autoregressive modeling.   
\vspace{0.1in}

\textbf{Advantages of the ECP Approach} - In the ECP approach, efficient attention computation occurs without any loss of contextual information down the layers of the Transformer. The ECP algorithm provides slightly better efficiency in terms of computations than the LongLoRA design due to PaerceiverAR attention. Figure~\ref{fig:LLP}(b-d) shows the sparse attention pattern as it is computed in the first layer, and how each subsequent layer increases the attention receptive field (because of overlap of half segments in the PerceiverAR i.e., PAR blocks). As shown in Figure~\ref{fig:LLP}(d), after enough layers, the information from all previous segments is available to the PAR block as it calculates the attention on the two consecutive half segments. Thus we achieve a similar benefit in calculating the attention as LongLoRA with slightly better efficiency, and more flexibility in the number of segments that the input sequence can be partitioned into. Note that the partitioning in LongLoRA is constrained by the number of heads.

One reason our ECP architecture performs better than full attention is that, each layer in the early stages is learning to predict the next token with less information. For example, in Figure~\ref{fig:LLPExp} the predicted tokens in $S_{3}^{1}$ (layer 1) only use the information from $S_{2}^{0}$ and $S_{3}^{0}$ and do not use context from $S_{0}^{0}$ and $S_{1}^{0}$. In layer 2, the computation of $S_{3}^{2}$ has information available from $S_{0}^{1}$, $S_{0}^{2}$ and $S_{0}^{3}$ (but not $S_{0}^{0}$). The following layer (layer 3) will have the information from all previous segments in computation of $S_{3}^{3}$. Thus, there is no loss of information in the next token prediction of any given segment in later layers, but initial layers are learning to predict the next tokens with less information. This is analogous to an implicit dropout in the earlier layers. This may be a contributing factor as to why our ECP model outperforms an equivalent size full attention Transformer.

In summary, the ECP approach combines efficiency and enhanced learning which results in a significantly improved Transformer architecture, as evidenced by the empirical results which we present next.

\section{Experimental Results}
\label{sec:experiments}

\begin{table*}
\small
\centering
\begin{tabular}{l|cccc|}
\cline{2-5}
                                  & \multicolumn{4}{c|}{Configuration A}                                                                                                                  \\ \cline{2-5} 
                                  & \multicolumn{1}{l|}{Baseline PerceiverAR} & \multicolumn{1}{l|}{Double Attention} & \multicolumn{1}{l|}{$s$-Split Double Attention} & \multicolumn{1}{l|}{Compressed Double Attention} \\ \hline
\multicolumn{1}{|l|}{Latent=256}  & \multicolumn{1}{c|}{63.8524}  & \multicolumn{1}{c|}{54.8534}          & \multicolumn{1}{c|}{55.4036}          & 63.112                                \\ \hline
\multicolumn{1}{|l|}{Latent=512}  & \multicolumn{1}{c|}{43.2374}  & \multicolumn{1}{c|}{38.6215}          & \multicolumn{1}{c|}{38.7562}          & 43.0051                               \\ \hline
\multicolumn{1}{|l|}{Latent=768}  & \multicolumn{1}{c|}{35.4916}  & \multicolumn{1}{c|}{32.9618}          & \multicolumn{1}{c|}{32.9865}          & 33.4186                               \\ \hline
                                  & \multicolumn{4}{c|}{Configuration B}                                                                                                                  \\ \cline{2-5} 
                                  & \multicolumn{1}{l|}{Baseline PerceiverAR} & \multicolumn{1}{l|}{Double Attention} & \multicolumn{1}{l|}{$s$-Split Double Attention} & \multicolumn{1}{l|}{Compressed Double Attention} \\ \hline
\multicolumn{1}{|l|}{Latent=1024} & \multicolumn{1}{c|}{31.902}   & \multicolumn{1}{c|}{30.3801}          & \multicolumn{1}{c|}{29.5097}          & 29.9752                               \\ \hline
                                  & \multicolumn{4}{c|}{Configuration C}                                                                                                                  \\ \cline{2-5} 
                                  & \multicolumn{1}{l|}{Baseline PerceiverAR} & \multicolumn{1}{l|}{Double Attention} & \multicolumn{1}{l|}{$s$-Split Double Attention} & \multicolumn{1}{l|}{Compressed Double Attention} \\ \hline
\multicolumn{1}{|l|}{Latent=1024} & \multicolumn{1}{c|}{28.2436}  & \multicolumn{1}{c|}{27.3821}          & \multicolumn{1}{c|}{27.1041}          & 26.713                                \\ \hline
\end{tabular}
\caption{Perplexity results for different architectures on the Wikitext-103 dataset. Configuration A, B and C represent different architecture variations and are fully detailed in Section~\ref{sec:experiments}.}
\label{table:MainResult}
\vspace{0.1in}
\end{table*}

\begin{table*}[t]
\small
\begin{tabular}{|>{\centering\arraybackslash}p{0.17\textwidth}|>{\centering\arraybackslash}p{0.14\textwidth}|>{\centering\arraybackslash}p{0.14\textwidth}|>{\centering\arraybackslash}p{0.14\textwidth}|>{\centering\arraybackslash}p{0.2\textwidth}|}
\hline
Model   Architecture & ECP - 12 Layers & ECP - 18 Layers & ECP - 24 Layers & PerceiverAR*  60 Layers \\ \hline
Model Size                     & 129.60M & 172.12M & 214.64M & 974.6M \\ \hline
Perplexity Wikitext-103 & 19.92          & 17.82 &\textbf{17.43}          & 18.35         \\ \hline
Perplexity PG-19        & 21.89          & 20.42 & \textbf{18.83}          & 28.9          \\ \hline
\end{tabular}
\caption{Test Perplexity results on Wikitext-103 and PG-19. All ECP models use 6 heads, an embedding dimension of 768, a sequence length of 2048 and a segment size of 256. PerceiverAR* indicates result cited from~\cite{hawthorne2022general}. ECP is our model.}
\label{table:MainResult3}
\vspace{0.1in}
\end{table*}

We evaluate the three proposed enhanced PerceiverAR architectures and the ECP architecture on Wikitext-103~\cite{merity2022pointer} and PG-19~\cite{raecompressive2019} datasets to test the perplexity of different models. In general, each model is trained under the same set of hyper-parameters (batch size, total number of epochs and learning rate scheduler) to make the perplexity comparisons standardized.

\vspace{0.1in}
\textbf{Results for Double Attention, Compressed Double Attention and $s$-Split Double Attention} - In Table~\ref{table:MainResult}, the main experimental results are presented for three different model configurations. For all configurations, the history length in the Perceiver model is the sequence length subtracted from the latent length. In Configuration A, a sequence length of 1024 is used with an embedding size of 512, with 8 heads and 8 layers. The $s$-Split Double Attention PerceiverAR uses a segment size of $s=256$ with latent sizes of 256 and 512. It uses a segment size of 128 when the latent length is 768. The Compressed Double Attention PerceiverAR compresses the history component to size 128. In Configuration B, the embedding size is 768, with 6 heads per layer and 6 layers. In this configuration, a sequence length of 2048 is used, with a latent length of 1024. The $s$-Split Double Attention PerceiverAR uses a segment size of 512. In Configuration A and B, each model is trained for 200,000 iterations with a starting learning rate of $2\times10^{-4}$. Lastly, in Configuration C, an embedding size of 768 is used and there are 6 heads per layer with a total of 6 layers. A sequence length of 2048 and a latent length of 1024 is used. The $s$-Split Double Attention PerceiverAR uses a segment size of 512 and the Compressed Double Attention PerceiverAR uses a compression size of 256. Models in this configuration are trained for 500,000 iterations.

From Table~\ref{table:MainResult}, it can be seen that the Double Attention PerceiverAR performs better than other PerceiverAR architecture paradigms, and significantly better than the baseline PerceiverAR. In the baseline model, the history information is absorbed into the latent part after the first layer. In contrast, the Double Attention PerceiverAR carries the history component to all layers, and keeps refining this information in each layer. The $s$-Split Double Attention PerceiverAR comes close to Double Attention’s performance, as it also carries the history information to all layers. However, in the $s$-Split Double Attention, the history is divided into disjoint segments, and the attention is only done within each segment. This results in more efficient computation, with some loss of information due to disjoint history segments. The Compressed Double Attention does not seem to improve the perplexity (lower is better) significantly as compared to the baseline. We present the results and analyses for our best performing ECP model next.

\vspace{0.1in}
\textbf{Results for Efficient Context Propagating Perceiver - ECP } - In Tables~\ref{table:MainResult2} and~\ref{table:MainResult3}, we present results of the ECP models on the Wikitext-103 and PG-19 datasets. ECP achieves better perplexity than the PercieverAR baseline with significantly less computations. This can be attributed to the pairwise overlapping application of the PerceiverAR style computation. This results in appropriate information extraction, and propagation of information of the entire context, down the layers of the transformer. Tables~\ref{table:litResults2} and~\ref{table:litResults} provide a comparison of our ECP model with other SOTA models, including the PercieverAR baseline model, on both the Wikitext-103 and PG-19 datasets. As can be seen from the results, our model achieves the lowest perplexity with the smallest model size (lower perplexity is better). 

We test our ECP architecture on the image part of the Long Range Arena (LRA) benchmark~\cite{taylong}. In the image part of LRA, the CIFAR-10 dataset (referred to as sCIFAR-10) is treated as a sequence of grayscale pixels. Table~\ref{table:scifar} presents the results on image classification for different Transformer architectures. The Transformer architectures compared in~\cite{taylong} used a smaller architecture with 3 layers and 4 heads in each layer. The embedding size used is 64, with a feedforward network size of 128. For fair architecture comparison, we also used the same sizes for one of our ECP models. Our Transformer implementation uses Rotary Position Embedding (RoPE~\cite{su2024roformer}) which yields better accuracy. Even though the state space models perform better than Transformer-based designs on sCIFAR-10, our ECP Transformer model produces the best known accuracy for Transformer-based architectures on the sCIFAR-10 dataset.

\vspace{0.1in}
\textbf{Efficiency and Complexity Analysis} -
The number of
calculation steps needed in the attention for different models is shown in Table \ref{table:ComplexityAnalysis}. Each attention step indicates calculation of
one attention entry. For example, in the canonical Transformer, the number of calculation steps in attention will be n × n.
As a comparison, the attention steps for different architectures are shown in Table \ref{table:ComplexityResults} with sequence length n = 4096, layers
l = 48, heads h = 24, segment size s = 256 and projection p = 256 .
As can be seen from Table 8, the ECP model is extremely efficient with only 12\% of the computation needed in attention
with respect to full attention in a Transformer. It also performs the best due to its pairwise extraction of attention information
and propagation of attention down the layers similar to Long LoRA.

\begin{table}
\small
\begin{tabular}{|l|l|l|}
\hline
Model   Architecture        & Model Size     & Perplexity      \\ \hline
\rowcolor{gray!25}
ECP (ours)                & 172.12 million & \textbf{17.82}  \\ \hline
\rowcolor{gray!25}
ECP (ours)                & 87 million     & \textbf{20.00} \\ \hline
xLSTM{[}7:1{]}$^{\ddag}$              & 163.7 million  & 21.47           \\ \hline
RWKV-4$^{\ddag}$                       & 169.4 million  & 22.33           \\ \hline
Mamba$^{\ddag}$                        & 167.8 million  & 22.49           \\ \hline
Llama$^{\ddag}$                        & 162.2 million  & 23.16           \\ \hline
H3 (Hungry Hungry Hippos)$^{\Delta}$ & 125 million    & 23.70            \\ \hline
Transformer-XL$^{\dagger}$              & 151 million    & 24.00              \\ \hline
$\infty$-Former**                    & 150 million    & 24.22           \\ \hline
\end{tabular}
\caption{Comparison of perplexity results on Wikitext-103 with published architectures of similar model sizes. $\dagger$ is from~\cite{dai-etal-2019-transformer}, $^{\ddag}$ is from~\cite{beck2024xlstm}, ** is from~\cite{martins2021infty} and $^\Delta$ is from~\cite{fu2022hungry}.}
\label{table:litResults2}
\end{table}

\begin{table}
\small
\begin{tabular}{|l|l|l|}
\hline
Model Architecture            & Model Size                 & Perplexity \\ \hline
\rowcolor{gray!25}
ECP Transformer (ours)                    &24L (214M)   & \textbf{18.83}      \\ \hline
Compressive   Transformer*     &36L (unknown) & 33.6       \\ \hline
Routing   Transformer*         & 22L (490M)  & 33.2       \\ \hline
Transformer-XL$^{\dagger}$                & 36L (unknown) & 36.3       \\ \hline
Block   Recurrent Transformer* & 24L (1.3B)  & 26.5       \\ \hline
\end{tabular}
\caption{Comparison of perplexity results for the ECP model and other models on PG-19 Dataset. * indicates results reported from~\cite{hutchins2022block} and $\dagger$ indicates results reported from~\cite{dai-etal-2019-transformer}. L=Number of layers and () gives the number of model parameters in either millions (M) or billions (B).}
\label{table:litResults}
\end{table}

\begin{table}
\centering
\begin{tabular}{|l|c|}
\hline
Model   Architecture & Test Accuracy   \\ \hline
\rowcolor{gray!25}
ECP (ours) – 12 layers        & \textbf{64.42\%}         \\ \hline
\rowcolor{gray!25}
ECP (ours) – 3 layers        & \textbf{59.32\%}         \\ \hline
Transformer   (with RoPE)                  & 51.32\%         \\ \hline
Transformer$^{***}$                                 & 42.44\%         \\ \hline
Sparse   Transformer$^{***}$                        & 44.24\%         \\ \hline
Performer$^{***}$                                   & 42.77\%         \\ \hline
Longformer$^{***}$                                  & 42.22\%         \\ \hline
Big Bird$^{***}$                                   & 40.83\%         \\ \hline
\end{tabular}
\caption{Comparison of Transformer architectures on the sCIFAR-10 dataset. $^{***}$ indicates results from~\cite{taylong}\label{table:scifar}}
\end{table}

\begin{table}
\small
\centering
\begin{tabular}{l|ccc|}
\cline{2-4}
                                           & \multicolumn{3}{c|}{ECP Architecture}                                                                   \\ \cline{2-4} 
                                           & \multicolumn{1}{c|}{S=512} & \multicolumn{1}{c|}{S=256} & S=128 \\ \hline
\multicolumn{1}{|l|}{SL=1024} & \multicolumn{1}{c|}{\textbf{20.2716}}          & \multicolumn{1}{c|}{20.5148}          & 20.7883          \\ \hline
\multicolumn{1}{|l|}{SL=2048} & \multicolumn{1}{c|}{\textbf{20.0021}}          & \multicolumn{1}{c|}{20.4536}          & 20.7482          \\ \hline
\end{tabular}
\caption{Perplexity results for the ECP model with different segment sizes on Wikitext-103. S= Segment Size. SL= Sequence Length.}
\label{table:MainResult2}
\end{table}

\begin{table*}[htbp]
\centering
\renewcommand{\arraystretch}{1.8}
\begin{tabular}{|>{\centering\arraybackslash}m{2.8cm}|>{\centering\arraybackslash}m{5.8cm}|}
\hline
\textbf{Model} & \textbf{Number of Calculation Steps in Attention} \\
\hline
PerceiverAR (baseline) & 
$\left\lceil \frac{n}{2} \times n \right\rceil + (l - 1) \left\lceil \frac{n}{2} \right\rceil^2 \times h$ \\
\hline
Double Attention & 
$\left[\left\lceil \frac{n}{2} \times n \right\rceil + \left\lceil \frac{n}{2} \times \frac{n}{2} \right\rceil\right] \times h \times l$ \\
\hline
Compressed Attention & 
$\left[\left\lceil \frac{n}{2s} \times \frac{n}{2s} \right\rceil \times s + \left\lceil \frac{n}{2} \times n \right\rceil\right] \times h \times l$ \\
\hline
$s$-Split Double Attention & 
$\left[\left\lceil \frac{n}{2} \times n \right\rceil + \left\lceil \frac{n}{2} \times \left(p + \frac{n}{2} \right) \right\rceil\right] \times (l - 1) + \left\lceil p \times p \right\rceil \times (l - 1) \times h$ \\
\hline
ECP Model  & 
$\left[\left\lceil \frac{s}{2} \times \frac{s}{2} \right\rceil + \left\lceil \frac{s}{2} \times s \right\rceil \times \left(\left\lceil \frac{n}{s/2} \right\rceil - 1\right)\right] \times h \times l$ \\
\hline
Regular Transformer & 
$(n \times n) \times h \times l$ \\
\hline
\end{tabular}
\caption{Attention Complexity in Different Models, n = sequence length, p = projection size, s = segment size, l = layers, h = heads}
\label{table:ComplexityAnalysis}
\end{table*}

\begin{table*}[ht] 
\centering 
\label{tab:model_calculations} 

\begin{tabular}{|l|r|r|}
\hline 
\textbf{Model} & \textbf{\shortstack{Number of Calculation \\ Steps in Attention}} & \textbf{\shortstack{Percent of Full \\ Attention}} \\
\hline 
PerceiverAR (baseline) & 4932 million & 25\% \\
\hline 
Double Attention & 14495 million & 75\% \\
\hline 
Compressed Double Attention & 9682 million & 50\% \\
\hline 
$s$-Split Double Attention & 14858 million & 77\% \\
\hline 
ECP Attention  & 2452 million & 12\% \\
\hline 
Transformer (full attention) & 19327 million & 100\% \\
\hline 
\end{tabular}
\caption{Relative computation efficiency of attention for different models with 24 heads and 48
layers. Sequence Length = 4096.}
\label{table:ComplexityResults}
\vspace{0.1in}
\end{table*}

\section{Conclusion}
\label{sec:Conclusion}
Efficient computation of attention in Transformer models is an important area of research, with significant impact on the design of LLMs. We start with the PerceiverAR architecture as a basis for developing enhanced architectures. PerceiverAR divides the input into two components i.e., the history and the latent. This two level breakdown can be exploited in different ways. We first propose three  distinct enhanced PerceiverAR architectural paradigms, each with their own efficiency performance trade-offs. Based on the two main limitations of the PerceiverAR (limited latent training and lossy history), and the strengths of different enhanced designs, we develop the Efficient Context Propagating Perceiver (ECP). The ECP architecture utilizes both the context and the latent sequences in autoregressive training. In addition, the ECP design operates with similar attention complexity as LongLoRA, making it extremely computationally efficient. It also implicitly extracts better information than existing architectures, via its overlapping pairwise segment attention.



\bibliography{mybibfile}

@article{su2024roformer,
  title={Roformer: Enhanced transformer with rotary position embedding},
  author={Su, Jianlin and Ahmed, Murtadha and Lu, Yu and Pan, Shengfeng and Bo, Wen and Liu, Yunfeng},
  journal={Neurocomputing},
  volume={568},
  pages={127063},
  year={2024},
  publisher={Elsevier}
}

@article{taylong,
  title={Long range arena: A benchmark for efficient transformers},
  author={Tay, Yi and Dehghani, Mostafa and Abnar, Samira and Shen, Yikang and Bahri, Dara and Pham, Philip and Rao, Jinfeng and Yang, Liu and Ruder, Sebastian and Metzler, Donald},
  journal={arXiv preprint arXiv:2011.04006},
  year={2020}
}

@article{fu2022hungry,
  title={Hungry hungry hippos: Towards language modeling with state space models},
  author={Fu, Daniel Y and Dao, Tri and Saab, Khaled K and Thomas, Armin W and Rudra, Atri and R{\'e}, Christopher},
  journal={arXiv preprint arXiv:2212.14052},
  year={2022}
}

@article{martins2021infty,
  title={$\infty$-former: Infinite Memory Transformer},
  author={Martins, Pedro Henrique and Marinho, Zita and Martins, Andr{\'e} FT},
  journal={arXiv preprint arXiv:2109.00301},
  year={2021}
}

@article{beck2024xlstm,
  title={xLSTM: Extended Long Short-Term Memory},
  author={Beck, Maximilian and P{\"o}ppel, Korbinian and Spanring, Markus and Auer, Andreas and Prudnikova, Oleksandra and Kopp, Michael and Klambauer, G{\"u}nter and Brandstetter, Johannes and Hochreiter, Sepp},
  journal={arXiv preprint arXiv:2405.04517},
  year={2024}
}

@article{hutchins2022block,
  title={Block-recurrent transformers},
  author={Hutchins, DeLesley and Schlag, Imanol and Wu, Yuhuai and Dyer, Ethan and Neyshabur, Behnam},
  journal={Advances in neural information processing systems},
  volume={35},
  pages={33248--33261},
  year={2022}
}

@article{raecompressive2019,
author = {Rae, Jack W and Potapenko, Anna and Jayakumar, Siddhant M and
          Hillier, Chloe and Lillicrap, Timothy P},
title = {Compressive Transformers for Long-Range Sequence Modelling},
journal = {arXiv preprint},
url = {https://arxiv.org/abs/1911.05507},
year = {2019},
}

@inproceedings{merity2022pointer,
  title={Pointer Sentinel Mixture Models},
  author={Merity, Stephen and Xiong, Caiming and Bradbury, James and Socher, Richard},
  booktitle={International Conference on Learning Representations},
  year={2022}
}

@article{vaswani2017attention,
  title={Attention is all you need},
  author={Vaswani, A},
  journal={Advances in Neural Information Processing Systems},
  year={2017}
}

@article{achiam2023gpt,
  title={Gpt-4 technical report},
  author={Achiam, Josh and Adler, Steven and Agarwal, Sandhini and Ahmad, Lama and Akkaya, Ilge and Aleman, Florencia Leoni and Almeida, Diogo and Altenschmidt, Janko and Altman, Sam and Anadkat, Shyamal and others},
  journal={arXiv preprint arXiv:2303.08774},
  year={2023}
}

@article{team2023gemini,
  title={Gemini: a family of highly capable multimodal models},
  author={Team, Gemini and Anil, Rohan and Borgeaud, Sebastian and Wu, Yonghui and Alayrac, Jean-Baptiste and Yu, Jiahui and Soricut, Radu and Schalkwyk, Johan and Dai, Andrew M and Hauth, Anja and others},
  journal={arXiv preprint arXiv:2312.11805},
  year={2023}
}

@article{touvron2023llama,
  title={Llama: Open and efficient foundation language models},
  author={Touvron, Hugo and Lavril, Thibaut and Izacard, Gautier and Martinet, Xavier and Lachaux, Marie-Anne and Lacroix, Timoth{\'e}e and Rozi{\`e}re, Baptiste and Goyal, Naman and Hambro, Eric and Azhar, Faisal and others},
  journal={arXiv preprint arXiv:2302.13971},
  year={2023}
}

@article{dubey2024llama,
  title={The llama 3 herd of models},
  author={Dubey, Abhimanyu and Jauhri, Abhinav and Pandey, Abhinav and Kadian, Abhishek and Al-Dahle, Ahmad and Letman, Aiesha and Mathur, Akhil and Schelten, Alan and Yang, Amy and Fan, Angela and others},
  journal={arXiv preprint arXiv:2407.21783},
  year={2024}
}

@inproceedings{dai-etal-2019-transformer,
    title = "Transformer-{XL}: Attentive Language Models beyond a Fixed-Length Context",
    author = "Dai, Zihang  and
      Yang, Zhilin  and
      Yang, Yiming  and
      Carbonell, Jaime  and
      Le, Quoc  and
      Salakhutdinov, Ruslan",
    editor = "Korhonen, Anna  and
      Traum, David  and
      M{\`a}rquez, Llu{\'\i}s",
    booktitle = "Proceedings of the 57th Annual Meeting of the Association for Computational Linguistics",
    month = jul,
    year = "2019",
    address = "Florence, Italy",
    publisher = "Association for Computational Linguistics",
    url = "https://aclanthology.org/P19-1285",
    doi = "10.18653/v1/P19-1285",
    pages = "2978--2988",
}

@misc{wang2020linformer,
    title={Linformer: Self-Attention with Linear Complexity},
    author={Sinong Wang and Belinda Z. Li and Madian Khabsa and Han Fang and Hao Ma},
    year={2020},
    eprint={2006.04768},
    archivePrefix={arXiv},
    primaryClass={cs.LG}
}

@article{beltagy2020longformer,
  title={Longformer: The long-document transformer},
  author={Beltagy, Iz and Peters, Matthew E and Cohan, Arman},
  journal={arXiv preprint arXiv:2004.05150},
  year={2020}
}

@inproceedings{kitaev2020reformer,
    title       = {Reformer: The Efficient Transformer},
    author      = {Nikita Kitaev and Lukasz Kaiser and Anselm Levskaya},
    booktitle   = {International Conference on Learning Representations},
    year        = {2020},
    url         = {https://openreview.net/forum?id=rkgNKkHtvB}
}

@inproceedings{choromanski2020rethinking,
  title={Rethinking Attention with Performers},
  author={Choromanski, Krzysztof Marcin and Likhosherstov, Valerii and Dohan, David and Song, Xingyou and Gane, Andreea and Sarlos, Tamas and Hawkins, Peter and Davis, Jared Quincy and Mohiuddin, Afroz and Kaiser, Lukasz and others},
  booktitle={International Conference on Learning Representations},
  year={2020}
}

@article{zhu2021long,
  title={Long-short transformer: Efficient transformers for language and vision},
  author={Zhu, Chen and Ping, Wei and Xiao, Chaowei and Shoeybi, Mohammad and Goldstein, Tom and Anandkumar, Anima and Catanzaro, Bryan},
  journal={Advances in neural information processing systems},
  volume={34},
  pages={17723--17736},
  year={2021}
}

@inproceedings{hawthorne2022general,
  title={General-purpose, long-context autoregressive modeling with Perceiver AR},
  author={Hawthorne, Curtis and Jaegle, Andrew and Cangea, C{\u{a}}t{\u{a}}lina and Borgeaud, Sebastian and Nash, Charlie and Malinowski, Mateusz and Dieleman, Sander and Vinyals, Oriol and Botvinick, Matthew and Simon, Ian and others},
  booktitle={International Conference on Machine Learning},
  pages={8535--8558},
  year={2022},
  organization={PMLR}
}

@inproceedings{jaegle2021perceiver,
  title={Perceiver: General perception with iterative attention},
  author={Jaegle, Andrew and Gimeno, Felix and Brock, Andy and Vinyals, Oriol and Zisserman, Andrew and Carreira, Joao},
  booktitle={International conference on machine learning},
  pages={4651--4664},
  year={2021},
  organization={PMLR}
}

@inproceedings{jaegleperceiver,
  title={Perceiver IO: A General Architecture for Structured Inputs \& Outputs},
  author={Jaegle, Andrew and Borgeaud, Sebastian and Alayrac, Jean-Baptiste and Doersch, Carl and Ionescu, Catalin and Ding, David and Koppula, Skanda and Zoran, Daniel and Brock, Andrew and Shelhamer, Evan and others},
  booktitle={International Conference on Learning Representations},
 year={2022}
}

@article{gu2021efficiently,
  title={Efficiently modeling long sequences with structured state spaces},
  author={Gu, Albert and Goel, Karan and R{\'e}, Christopher},
  journal={arXiv preprint arXiv:2111.00396},
  year={2021}
}

@inproceedings{fuhungry,
  title={Hungry Hungry Hippos: Towards Language Modeling with State Space Models},
  author={Dao, Tri and Fu, Daniel Y and Saab, Khaled K and Thomas, Armin W and Rudra, Atri and R{\'e}, Christopher},
  booktitle={Proceedings of the 11th International Conference on Learning Representations (ICLR)},
  year={2023}
}

@inproceedings{daotransformers,
  title={Transformers are SSMs: Generalized Models and Efficient Algorithms Through Structured State Space Duality},
  author={Dao, Tri and Gu, Albert},
  booktitle={Forty-first International Conference on Machine Learning},
year={2024}
}

@article{chenlonglora,
  title={Longlora: Efficient fine-tuning of long-context large language models},
  author={Chen, Yukang and Qian, Shengju and Tang, Haotian and Lai, Xin and Liu, Zhijian and Han, Song and Jia, Jiaya},
  journal={arXiv preprint arXiv:2309.12307},
  year={2023}
}

\end{document}